\documentclass[conference]{IEEEtran}
\IEEEoverridecommandlockouts
\usepackage{cite}
\usepackage{amsmath,amssymb,amsfonts}
\usepackage{algorithmic}
\usepackage{graphicx}
\usepackage{textcomp}
\usepackage{xcolor}
\def\BibTeX{{\rm B\kern-.05em{\sc i\kern-.025em b}\kern-.08em
    T\kern-.1667em\lower.7ex\hbox{E}\kern-.125emX}}
\begin{document}

\title{Attention vs LSTM: Improving Word-level BISINDO Recognition}

\author{
\centering
\IEEEauthorblockN{1\textsuperscript{st} Muchammad Daniyal Kautsar}
\IEEEauthorblockA{\textit{Department of Electrical and} \\
\textit{Information Engineering}\\
\textit{Universitas Gadjah Mada}\\
Yogyakarta, Indonesia \\
muchammad.daniyal.kautsar@mail.ugm.ac.id}
\and
\IEEEauthorblockN{2\textsuperscript{nd} Afra Majida Hariono}
\IEEEauthorblockA{\textit{Department of Electrical and} \\
\textit{Information Engineering}\\
\textit{Universitas Gadjah Mada}\\
Yogyakarta, Indonesia \\
afra.majida0202@mail.ugm.ac.id}
\and
\centering
\IEEEauthorblockN{3\textsuperscript{rd} Ridwan Akmal}
\IEEEauthorblockA{\textit{Department of Electrical} \\
\textit{Engineering and Informatics}\\
\textit{Universitas Gadjah Mada}\\
Yogyakarta, Indonesia \\
ridwan.akmal0202@mail.ugm.ac.id}
}

\maketitle

\begin{abstract}
Indonesia ranks fourth globally in the number of deaf cases. Individuals with hearing impairments often find communication challenging, necessitating the use of sign language. However, there are limited public services that offer such inclusivity. On the other hand, advancements in artificial intelligence (AI) present promising solutions to overcome communication barriers faced by the deaf. This study aims to explore the application of AI in developing models for a simplified sign language translation app and dictionary, designed for integration into public service facilities, to facilitate communication for individuals with hearing impairments, thereby enhancing inclusivity in public services. The researchers compared the performance of LSTM and 1D CNN + Transformer (1DCNNTrans) models for sign language recognition. Through rigorous testing and validation, it was found that the LSTM model achieved an accuracy of 94.67\%, while the 1DCNNTrans model achieved an accuracy of 96.12\%. Model performance evaluation indicated that although the LSTM exhibited lower inference latency, it showed weaknesses in classifying classes with similar keypoints. In contrast, the 1DCNNTrans model demonstrated greater stability and higher F1 scores for classes with varying levels of complexity compared to the LSTM model. Both models showed excellent performance, exceeding 90\% validation accuracy and demonstrating rapid classification of 50 sign language gestures.
\end{abstract}


\section{Introduction}

Based on data from the Ministry of Health of the Republic of Indonesia in 2019, around 18.9 million or equivalent to 6.8\% of Indonesians experience mild or severe hearing loss, this makes deafness the fourth most prevalent disability case in Indonesia. Hearing impairment, commonly referred to as deafness, is a condition where an individual experiences partial or complete loss of hearing, making verbal communication challenging \cite{rahmah2018problematika}. Based on Law No. 8 of 2016 Article 26 concerning persons with disabilities, there are matters that discuss the right to be free from discrimination for persons with disabilities, which includes the right to socialize and interact in social and state life without fear. For this reason, a special language is applied so that communication can be carried out between deaf people, as well as deaf people with hearing people using sign language. Although they can see, not all deaf people can understand verbal language such as writing, for this reason the use of sign language is easier to use because it prioritizes visuals such as hand gestures, body, and facial expressions to communicate. In Indonesia, the sign language commonly used in daily life is Indonesian Sign Language (BISINDO) \cite{handhika2018gesture}.

Although it has been regulated in law, unfortunately there are not many public service facilities that employ employees with sign language skills, in the social community there are also not many people who learn and are able to use sign language. In fact, communication is the main aspect in building relationships and interactions between individuals. All social activities such as school, commerce, and public services require intense communication from both directions. With the ease of technology, to realize inclusivity in public services and other social activities, the author created a technology based on artificial intelligence (AI) with the main foundation in the form of deep learning and computer vision. This technology is made in the form of applications that can help support the resolution of these problems. Existing technology can be developed to build sign language detection technology in BISINDO based on deep learning. The application of this technology can be further developed in terms of translators and simple sign language dictionaries that can be applied in public service facilities, so that officers can easily communicate with deaf people. In addition, a learning system can also be developed that can expand the inclusiveness of public services.

\section{Methodology}
\subsection{Dataset Collection and Preprocessing}
In this study, the author collected data by recording videos consisting of 50 classes containing alphabetic letters and some important expressions, such as “tolong”, "halo”, “maaf”, and so on. In each class, 139 videos were collected with a duration of 50 frames for each video. Next, MediaPipe with BlazePose  model was used to perform keypoint detection on the body. First, the data is captured using a cellphone camera in the form of a video, then the MediaPipe holistic model will detect keypoints. The results of this detection will be represented in a landmark in each frame, then the data will be stored in a folder that has been defined various actions in the context of gestures such as “halo”, “aku”, “perkenalkan” and others. For each action in the video data will be collected and extracted into a numpy array. 

In addition to using our own dataset, the authors also use the Google Isolated Sign Language Recognition (ISLR) Corpus dataset \cite{islr-corpus}. This dataset contains American sign language keypoints consisting of 250 classes, there are a total of 94 thousand different data in this dataset. The authors used this dataset during the pre-training process of the model, to provide initial knowledge to the model.

\subsection{Model Architecture}
\subsubsection{Long-Short Term Memory (LSTM) Architecture}
Long-Short Term Memory or LSTM is one of the deep learning algorithms with a type of artificial neural network that is a development of Recurrent Neural Network (RNN) \cite{Hochreiter-lstm}. LSTM is a method that can learn and maintain long-term data information and perform sequential data processing, LSTM is suitable for use in sign language processing because gestures that represent words or phrases have complex long-term dependencies and will have different meanings if not done in order \cite{daniyal, sharma2022automated}. 

\begin{figure}[htbp]
\centerline{\includegraphics[width=0.95\linewidth]{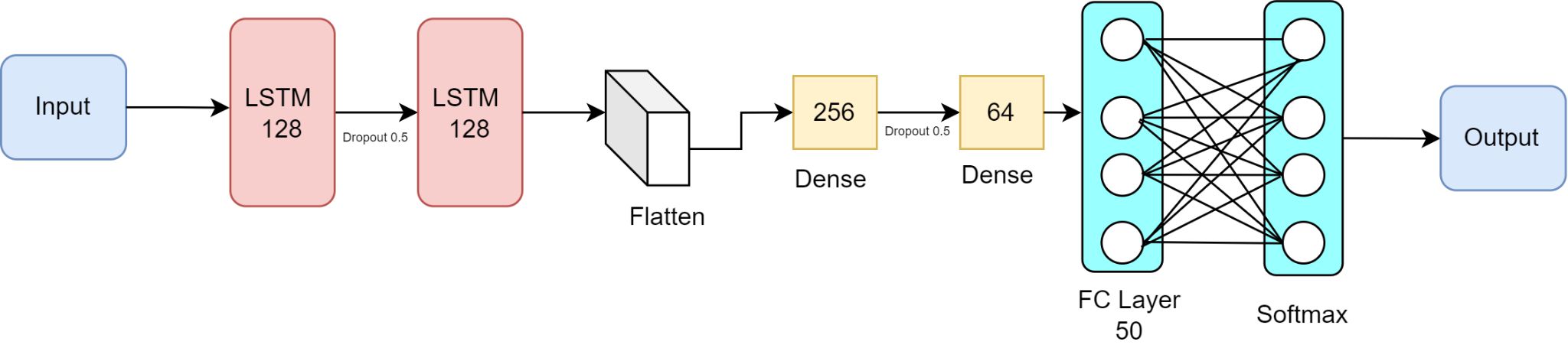}}
\caption{Proposed LSTM architecture.}
\label{fig3-1}
\end{figure}

The LSTM architecture consists of a 128-unit LSTM layer, as shown in Fig.~\ref{fig3-1}. This layer uses the ReLU activation function and receives input with a size of 45 x 174, the value 45 is the sequence length of the input and 174 is the number of features from each video frame that has been processed. After the first LSTM layer is passed, a dropout layer of 0.5 is given which will deactivate 50\% of the neurons randomly. After flattening and adding dense layers, the first layer with 256 units is passed again to the ReLU activation function with the same scheme as the previous LSTM layer, this layer will be given a dropout. Finally, after going through an iterative process between the LSTM and dense layer model, a softmax activation function is applied to classify the input into one of the 50 predefined motion classes.

\subsubsection{1-Dimensional CNN + Transformer (1DCNNTrans) Architecture}
Convolutional Neural Network or CNN is the development of ANN by improving its shift and translational invariance \cite{485891}. CNNs generally consist of a convolutional layer consisting of a filter (kernel) that traverses the input signal and produces a feature map. CNN blocks are usually composed of convolutional layers, pooling layers, and at the end there is a fully connected layer as a link from all activities in the previous layer \cite{kiranyaz20191, gama2018convolutional}. 

On the other hand, Transformer is an architecture that relies on self-attention to calculate input and output representations without using RNN or convolutional layers, this model is commonly used to translate sign language \cite{vaswani2017attention}. The transformer architecture consists of an encoder stack composed of multi-head self-attention and feed-forward position-wise layers. In addition, there is a decoder to generate an output sequence based on the final representation of the encoder, and a fully connected layer to linearly transform the representation generated by the encoder \cite{chaudhary2022signnet}.

\begin{figure}[htbp]
\centerline{\includegraphics[width=0.95\linewidth]{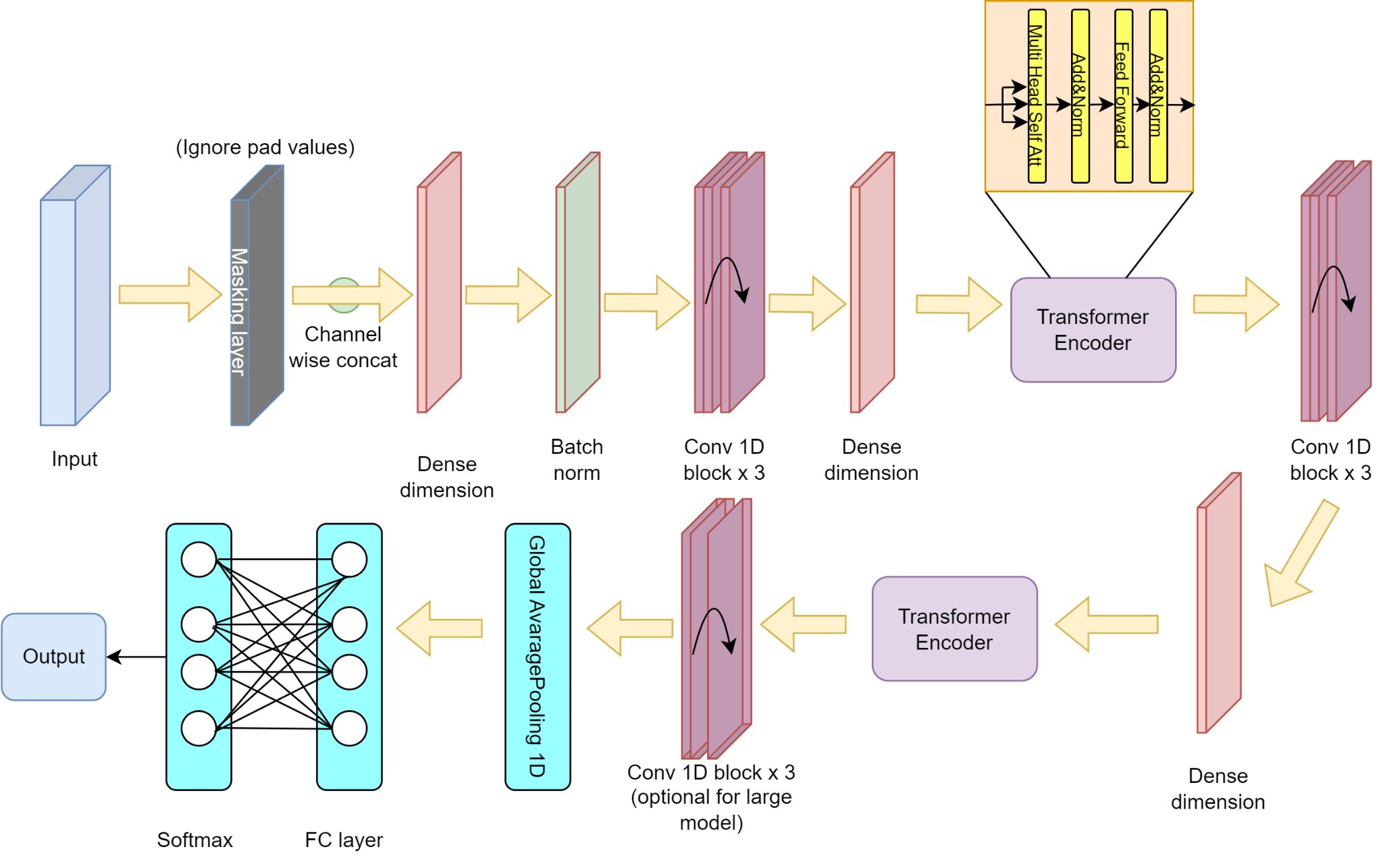}}
\caption{Proposed 1D CNN + Transformer (1DCNNTrans) architecture.}
\label{fig3-2}
\end{figure}

The CNN architecture in this study is built using one-dimensional convolutional blocks. It first initializes the maximum length and number of channels used when feeding the input into the first layer. The input will be forwarded to masking, a layer that is useful for handling sequences of varying lengths. Next, to enrich the features, the input channels will be given an expand ratio factor in the dense layer, this layer will return the number of channels to the original size that has been determined so that they can be processed further into the transformer block that will be built. During the process that occurs in the convolutional block, batch normalization is added to keep the activation distribution stable and use a residual connection scheme to accelerate convergence during the training phase. The overall architecture of the combination between 1D CNN and block transformer, which we have named 1DCNNTrans, can be seen in Fig.~\ref{fig3-2}.

\subsubsection{Efficient Channel Attention (ECA)}
\begin{figure}[htbp]
\centerline{\includegraphics[width=0.95\linewidth]{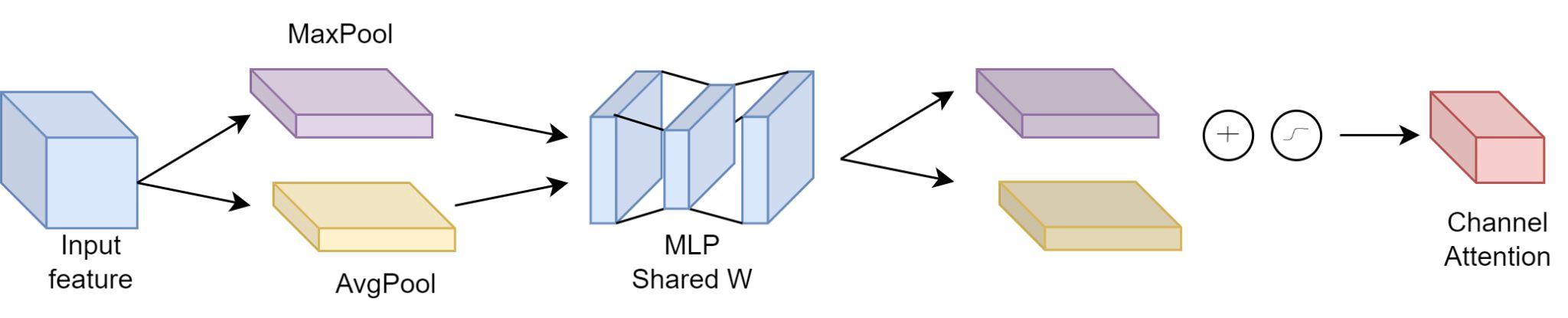}}
\caption{Efficient Channel Attention (ECA) architecture.}
\label{fig3-3}
\end{figure}

\begin{figure}[htbp]
\centerline{\includegraphics[width=0.6\linewidth]{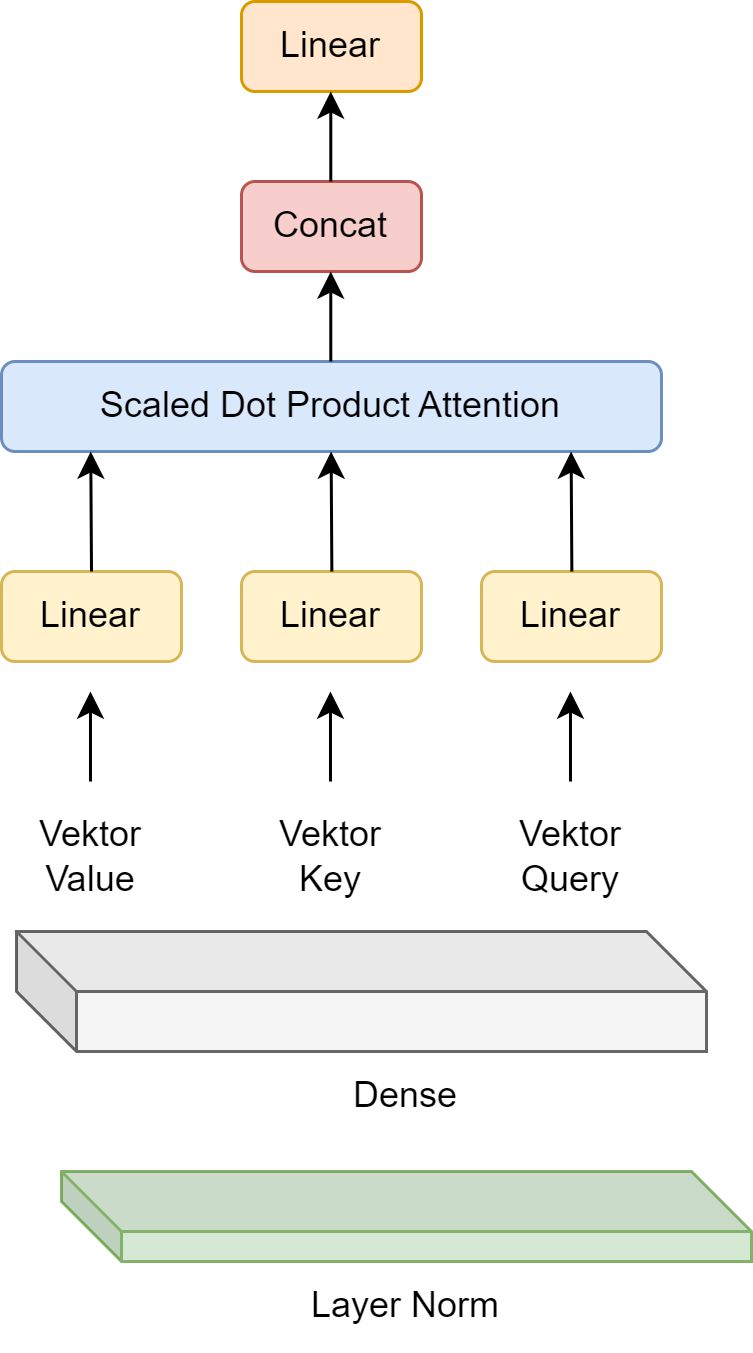}}
\caption{Multi-head Self-attention architecture.}
\label{fig3-4}
\end{figure}

In each convolutional block constructed, Efficient Channel Attention (ECA) is applied. ECA works by calculating the max pooling and global average pooling of the input, and then generating a channel attention scale through a convolutional layer with a small kernel. This scale is then applied to the original input to highlight the more important features. ECA has the advantage of not requiring many additional parameters, but is still effective in improving network performance by focusing on important information between channels \cite{wang2020eca}. By using a small kernel to determine the attention weight without requiring a fully connected layer, the layer finally succeeds in generating an attention scale valued between 0 and 1 that is given a sigmoid function to amplify important features. An illustration of ECA can be seen in Fig.~\ref{fig3-3}. After the number of channels returns to the original size, the data will be processed to the transformer block. Before entering the multi-head self-attention layer in Fig.~\ref{fig3-4}, the numerical input will be stabilized with the normalization layer. 

Next with the dense layer, the input will produce three different vectors namely query(q), key(k), and value(v). The attention calculation is done by multiplying the transposed query(q) and key(k) vectors and then adding the scaling. From these results, a softmax function is applied that can provide a decision on which input will be given more attention. The results obtained will be projected back to the original dimension through the dense layer. During the process, a residual connection (add) process will be performed afterward. The output of the multi-head self-attention will be added directly to the original input of this block which will continue with the norm process using the normalization layer which is useful in stabilizing the gradient. Finally, the feed-forward network will be initialized with two dense layers that utilize activation functions to further process information and perform non-linear transformation of the feature representation.

\section{Experiments}
\subsection{Hyperparameter tuning}

In this research, two approaches are used namely 1DCNNTrans architecture and LSTM architecture. The hyperparameters used for each architecture have been optimized to achieve the best performance. Table.~\ref{tab3-1} shows the hyperparameter configuration applied to the 1DCNNTrans and to the LSTM architecture built.

\begin{table}[htbp]
\caption{Hyperparameters}
\begin{center}
\renewcommand{\arraystretch}{1.75}
\begin{tabular}{p{5cm} p{3cm}}
\hline
\textbf{Hyperparameter} & \textbf{Configuration} \\ \hline
n splits & 5 \\ 
seed & 42 \\ 
max length & 384 \\ 
learning rate (start for optimizer) & 0.00005 \\ 
learning rate (minimum when decay) & 0.000001 \\ 
weight decay & 0.1 \\ 
batch size & 64 \\ 
decay type & cosine \\ \hline
\end{tabular}
\renewcommand{\arraystretch}{1}  
\label{tab3-1}
\end{center}
\end{table}

\subsection{Training}
The training phase is divided into two steps, the first is the pre-training process on the Google-ISLR Corpus dataset and the second by fine tuning the dataset that the author has collected. Both steps receive the same settings on hyperparameters, seed, dropout, and other training parameters. Furthermore, each training result will be forwarded to the two architectures that the author proposes in this study. For the LSTM architecture, the training phase will be trained with 200 iterations using a callback initialized benchmark of the loss to be achieved by stopping the training phase if the loss of the training phase is greater than the loss in the validation phase which indicates the potential for overfitting.

For the 1D CNN phase combined with transformers when the training phase begins, settings are made for several hyperparameters that have been set. The process starts with initializing the seed settings to ensure reproducibility. Datasets that have been divided into training data and validation data will be loaded for further configuration, then augmentation and other configurations are performed. The model is initialized with a function that will load the combination between the 1D CNN model and the transformer architecture that has been built previously. During this process, several parameters are also added, such as dropout step and scheduling process on the learning rate using OneCycleLR \cite{smith2019super} parameter that will modulate the learning rate value dynamically during the training process in one full cycle. The optimizer used is Rectified Adam, which is combined with Lookahead for training stability \cite{liu2020variance}. The model is then compiled with the optimizer, categorical cross entropy loss function, and Categorical Accuracy metric. Several callbacks are set up for logging (CSVLogger), model checkpointing (ModelCheckpoint), and stochastic weight averaging (SWA). Model training is done with training data, number of epochs, steps per epoch, callbacks, validation data, and verbosity that have been set in the previous configuration. After the training process is complete, the best model weight saved during training is loaded, and the model is evaluated on the validation data. By using a combination of 1D CNN and Transformer in the model architecture, the authors can utilize the advantages of both approaches to produce models that are more accurate and effective in understanding complex data and have diverse dimensions of both spatial and temporal structures.

\section{Results and Discussion}

\subsection{Pre-Training Results}
In the initial stage, the author conducted a pre-training process on the model using the Google-ISLR Corpus dataset. Each model is pre-trained for 50 epochs with a minimum accuracy target given by the author of 75\% of the 250 classes on the Google-ISLR Corpus dataset. This stage is carried out to provide initial knowledge to the model, so that the model can understand the dataset more easily and more accurately. The accuracy results achieved by each model are 76.82\% for the LSTM model and 81.94\% for the 1DCNNTrans model. 

\subsection{Training Results and Analysis}
The author trained on a dataset of 200 epochs for the LSTM model and 150 epochs for the 1DCNNTrans model. The dataset is split into 80\% training data and 20\% validation data. The accuracy results on the LSTM model reached 94.67\% on the validation split and the accuracy results on the 1DCNNTrans model amounted to 96.12\%. Accuracy results for validation data and average FPS can be seen in Table.~\ref{tab4-1}.

\begin{table}[htbp]
\caption{Model Performance}
\begin{center}
\renewcommand{\arraystretch}{1.75}
\begin{tabular}{p{2cm} p{3cm} p{2cm}}
\hline
\textbf{Model} & \textbf{Validation Accuracy}& \textbf{Average FPS$^{\mathrm{*}}$} \\
\hline
LSTM & 94.67\% & \textbf{102} \\
1DCNNTrans & \textbf{96.12\% }& 64 \\
\hline
\multicolumn{3}{l}{$^{\mathrm{*}}$Evaluate on Arm M1 CPU.}
\end{tabular}
\renewcommand{\arraystretch}{1}  
\label{tab4-1}
\end{center}
\end{table}

\begin{figure*}[htbp]
\centerline{\includegraphics[width=0.8\linewidth]{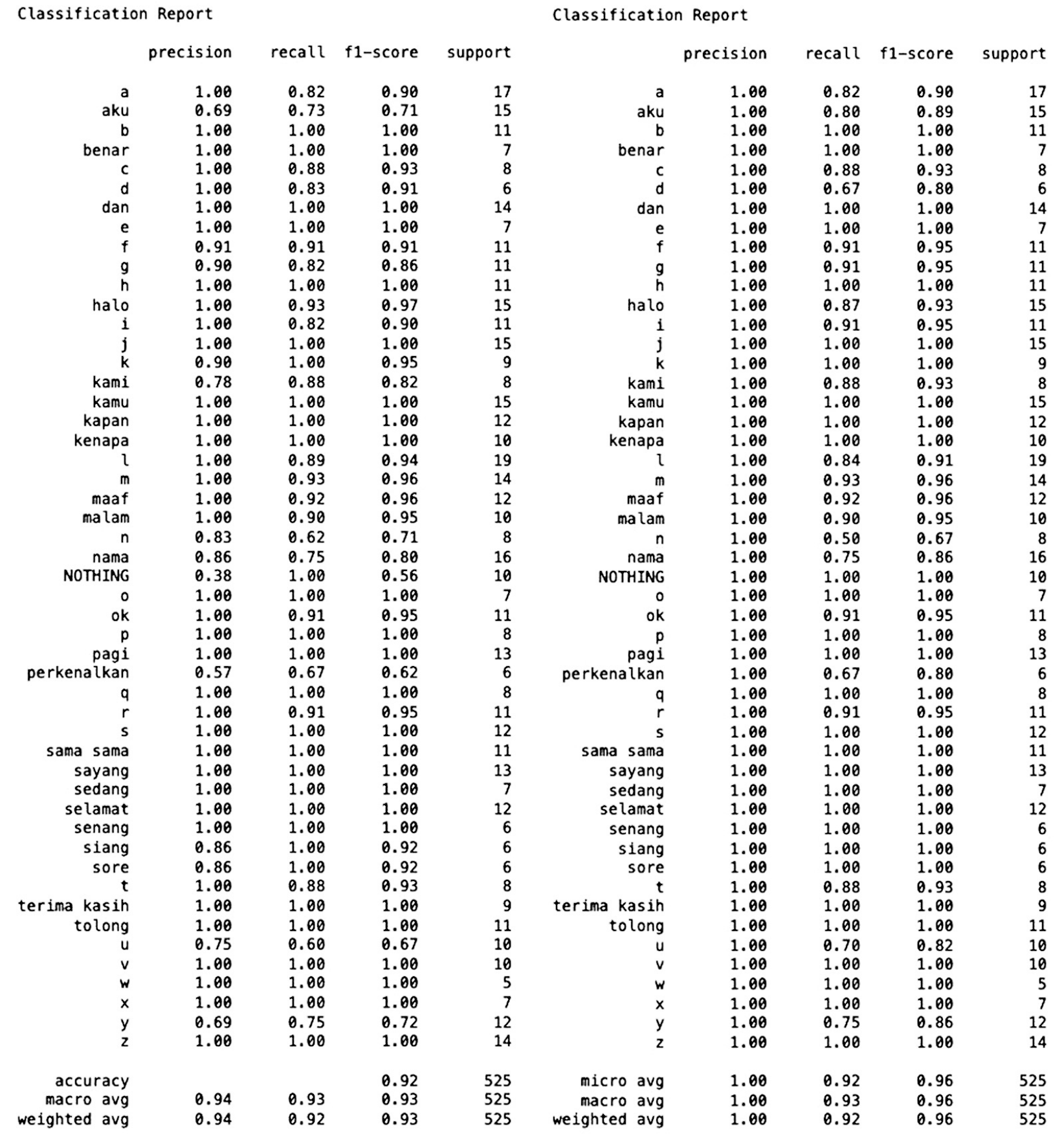}}
\caption{Classification Report. (Left) LSTM and (Right) 1DCNNTrans}
\label{fig4-1}
\end{figure*}

In both models, the author also compares the classification report to see the performance of the model in each class.  This comparison can be seen in Fig.~\ref{fig4-1}. In the LSTM model, it can be seen that this model has a fairly low performance in the class of movements that represent “i”, “u”, “n”, “perkenalkan”, “NOTHING”, and “y” with an f1-score value that is below 70\% compared to the 1DCNNTrans model which looks more stable and has a higher f1-score. Based on these performance results, it can be analyzed that the LSTM model has a weakness when dealing with classes with relatively similar keypoint sequences and complex movements. The author also considers that there is an anomaly in the results of the “NOTHING” class in LSTM which tends to be low, because the class should have a fairly low complexity. The “NOTHING” class is only composed of videos of people who do not perform sign language movements or transition processes when there is a pause in the conversation. The initial cause of this anomaly is because the “NOTHING” class movements are found at the beginning and end of the other classes. What is interesting here is that the performance of the 1DCNNTrans improves significantly on classes with poor performance on the LSTM. It can be seen that this model is much better at distinguishing motion sequences between classes, especially in classes with a certain level of complexity.

In Table.~\ref{tab4-1}, it can be seen that the LSTM model has a much lower latency compared to the 1DCNNTrans model. This lower latency is certainly more favorable towards its application in the real world, especially on devices with low specifications. Keep in mind that the model's inference latency has not gone through an optimization process with quantization, so the author cannot directly justify the possibility of performance in real-world scenarios, because when the optimization process is carried out on devices such as smartphones, there will be a possibility of different performance when the author conducts testing. Moreover, the longest inference latency is only at 64 fps, which certainly does not interfere with the normal fps duration of around 30 fps. But in general, both models performed very well with fairly fast inference. Both can classify 50 different classes with validation accuracy above 90\%.

\section{Conclusion}
Based on the research that has been done, there are four conclusions based on the research questions above, namely:
\begin{enumerate}

\item Artificial intelligence technology can be used to create a model that can later be used in making simple sign language translator and dictionary applications that can be applied in public service facilities, so that officers can easily communicate with deaf people. In addition, a learning system can also be developed that can expand the inclusiveness of public services.

\item Based on testing for split validation, the accuracy of the LSTM model reached 94.67\% and the accuracy of the 1DCNNTrans model was 96.12\%. 

\item The performance obtained from testing is that LSTM has a lower inference latency, the LSTM model shows weakness in classifying classes with similar keypoint sequences, especially in the “NOTHING” class which has an f1-score below 70\%. Meanwhile, the 1DCNNTrans model is more stable and has a higher f1-score on classes with a certain level of complexity compared to the LSTM model.

\item Both models performed very well with validation accuracy above 90\% and the ability to classify 50 classes of sign language gestures quickly. However, the combined 1D CNN and Transformer model is superior in handling classes with complex gestures and similar keypoint sequences, making it more effective in understanding complex data that has multiple dimensions in both spatial and temporal structure. However, the lower inference latency of the LSTM model provides a distinct advantage for application on lower-specification devices, making it still relevant in real-world scenarios.
\end{enumerate}

\bibliography{references}{}

\begin{thebibliography}{10}
\providecommand{\url}[1]{#1}
\csname url@samestyle\endcsname
\providecommand{\newblock}{\relax}
\providecommand{\bibinfo}[2]{#2}
\providecommand{\BIBentrySTDinterwordspacing}{\spaceskip=0pt\relax}
\providecommand{\BIBentryALTinterwordstretchfactor}{4}
\providecommand{\BIBentryALTinterwordspacing}{\spaceskip=\fontdimen2\font plus
\BIBentryALTinterwordstretchfactor\fontdimen3\font minus \fontdimen4\font\relax}
\providecommand{\BIBforeignlanguage}[2]{{%
\expandafter\ifx\csname l@#1\endcsname\relax
\typeout{** WARNING: IEEEtran.bst: No hyphenation pattern has been}%
\typeout{** loaded for the language `#1'. Using the pattern for}%
\typeout{** the default language instead.}%
\else
\language=\csname l@#1\endcsname
\fi
#2}}
\providecommand{\BIBdecl}{\relax}
\BIBdecl

\bibitem{rahmah2018problematika}
F.~N. Rahmah, ``Problematika anak tunarungu dan cara mengatasinya,'' \emph{Quality}, vol.~6, no.~1, pp. 1--15, 2018.

\bibitem{handhika2018gesture}
T.~Handhika, R.~Zen, D.~Lestari, I.~Sari \emph{et~al.}, ``Gesture recognition for indonesian sign language (bisindo),'' in \emph{Journal of Physics: Conference Series}, vol. 1028, no.~1.\hskip 1em plus 0.5em minus 0.4em\relax IOP Publishing, 2018, p. 012173.

\bibitem{islr-corpus}
\BIBentryALTinterwordspacing
M.~S. Ashley~Chow, Glenn~Cameron \emph{et~al.}, ``Google - isolated sign language recognition,'' 2023. [Online]. Available: \url{https://kaggle.com/competitions/asl-signs}
\BIBentrySTDinterwordspacing

\bibitem{Hochreiter-lstm}
\BIBentryALTinterwordspacing
S.~Hochreiter and J.~Schmidhuber, ``{Long Short-Term Memory},'' \emph{Neural Computation}, vol.~9, no.~8, pp. 1735--1780, 11 1997. [Online]. Available: \url{https://doi.org/10.1162/neco.1997.9.8.1735}
\BIBentrySTDinterwordspacing

\bibitem{daniyal}
\BIBentryALTinterwordspacing
{Abfertiawan, Muhammad Sonny}, {Kautsar, Muchammad Daniyal}, {Hasan, Faiz}, {Palinggi, Yoseph}, and {Pranoto, Kris}, ``The application of artificial neural network model to predicting the acid mine drainage from long-term lab scale kinetic test,'' \emph{E3S Web of Conf.}, vol. 485, p. 02012, 2024. [Online]. Available: \url{https://doi.org/10.1051/e3sconf/202448502012}
\BIBentrySTDinterwordspacing

\bibitem{sharma2022automated}
K.~Sharma, K.~A. Aaryan, U.~Dhangar, R.~Sharma, and S.~Taneja, ``Automated indian sign language recognition system using lstm models,'' in \emph{2022 International Conference on Computing, Communication, and Intelligent Systems (ICCCIS)}.\hskip 1em plus 0.5em minus 0.4em\relax IEEE, 2022, pp. 461--466.

\bibitem{485891}
A.~Jain, J.~Mao, and K.~Mohiuddin, ``Artificial neural networks: a tutorial,'' \emph{Computer}, vol.~29, no.~3, pp. 31--44, 1996.

\bibitem{kiranyaz20191}
S.~Kiranyaz, T.~Ince, O.~Abdeljaber, O.~Avci, and M.~Gabbouj, ``1-d convolutional neural networks for signal processing applications,'' in \emph{ICASSP 2019-2019 IEEE International Conference on Acoustics, Speech and Signal Processing (ICASSP)}.\hskip 1em plus 0.5em minus 0.4em\relax IEEE, 2019, pp. 8360--8364.

\bibitem{gama2018convolutional}
F.~Gama, A.~G. Marques, G.~Leus, and A.~Ribeiro, ``Convolutional neural network architectures for signals supported on graphs,'' \emph{IEEE Transactions on Signal Processing}, vol.~67, no.~4, pp. 1034--1049, 2018.

\bibitem{vaswani2017attention}
A.~Vaswani, ``Attention is all you need,'' \emph{Advances in Neural Information Processing Systems}, 2017.

\bibitem{chaudhary2022signnet}
L.~Chaudhary, T.~Ananthanarayana, E.~Hoq, and I.~Nwogu, ``Signnet ii: A transformer-based two-way sign language translation model,'' \emph{IEEE Transactions on Pattern Analysis and Machine Intelligence}, vol.~45, no.~11, pp. 12\,896--12\,907, 2022.

\bibitem{wang2020eca}
Q.~Wang, B.~Wu, P.~Zhu, P.~Li, W.~Zuo, and Q.~Hu, ``Eca-net: Efficient channel attention for deep convolutional neural networks,'' in \emph{Proceedings of the IEEE/CVF conference on computer vision and pattern recognition}, 2020, pp. 11\,534--11\,542.

\bibitem{smith2019super}
L.~N. Smith and N.~Topin, ``Super-convergence: Very fast training of neural networks using large learning rates,'' in \emph{Artificial intelligence and machine learning for multi-domain operations applications}, vol. 11006.\hskip 1em plus 0.5em minus 0.4em\relax SPIE, 2019, pp. 369--386.

\bibitem{liu2020variance}
L.~Liu, H.~Jiang, P.~He, W.~Chen, X.~Liu, J.~Gao, and J.~Han, ``On the variance of the adaptive learning rate and beyond,'' in \emph{8th International Conference on Learning Representations, ICLR 2020}, 2020.

\end{thebibliography}
\bibliographystyle{IEEEtran}

\end{document}